# A Dynamic Keypoints Selection Network for 6DoF Pose Estimation

Haowen Sun[1], Taiyong Wang[1,2]
1. School of Mechanical Engineering, Tianjin University, Tianjin 300350, China
2. Tianjin University Ren'ai College, Tianjin 301636, China

## Abstract

6 DoF poses estimation problem aims to estimate the rotation and translation parameters between two coordinates, such as object world coordinate and camera world coordinate. Although some advances are made with the help of deep learning, how to full use scene information is still a problem. Prior works tackle the problem by pixel-wise feature fusion but need to randomly selecte numerous points from images, which can not satisfy the demands of fast inference simultaneously and accurate pose estimation. In this work, we present a novel deep neural network based on dynamic keypoints selection designed for 6DoF pose estimation from a single RGBD image. Our network includes three parts, instance semantic segmentation, edge points detection and 6DoF pose estimation. Given an RGBD image, our network is trained to predict pixel category and the translation to edge points and center points. Then, a least-square fitting manner is applied to estimate the 6DoF pose parameters. Specifically, we propose a dynamic keypoints selection algorithm to choose keypoints from the foreground feature map. It allows us to leverage geometric and appearance information. During 6DoF pose estimation, we utilize the instance semantic segmentation result to filter out background points and only use foreground points to finish edge points detection and 6DoF pose estimation. Experiments on two commonly used 6DoF estimation benchmark datasets, YCB-Video and LineMoD, demonstrate that our method outperforms the state-of-the-art methods and achieves significant improvements over other same category methods time efficiency.

## 1. Introduction

Estimating 6DoF poses of objects from images is a necessary component in lots of real-world applications, such as robot manipulation [13, 16], autonomous navigation [14, 18, 26], and virtual reality [12]. It is hard to recognize the 3D location and orientation of an object in a canonical frame because of variations in lighting, sensor noise, occlusion of scenes and truncation of objects. Traditional approaches like [33, 38] used empirical hand-crafted features to extract the correspondence between object mesh models and images. Such hand-crafted local features would suffer from a restricted performance by scenes with heavy occlusion and changing illumination conditions. Recently, the explosive growth of deep learning techniques motivates several works to tackle this problem by convolution neural networks (CNNs) on RGB images [19, 25, 40] and reveal promising improvements. Although the RGB images provide abundant appearance information, the lack of scene geometry information caused by perspective projection can not show robustness towards challenging scenarios. The development of inexpensive RGBD sensors provides extra depth information to tackle the problem and also leads to a new challenge: How to make full use of the two scene information to get better pose estimation?

One line of existing works [19, 24, 26] leverage the advantage of the two data sources within pixel-wise feature fusion and directly regress the 6DoF pose parameters. These works like [21] apply a CNN and a point cloud network (PCN) to extract dense features from the RGB image and point cloud respectively and the extracted dense features are then naive concatenated for directly regressing the translation parameters $t$ and the quaternion rotation $R$. However, due to the nonlinearity of the rotation space illustrated by [25], these methods typically have poor generalization. On the other hand, works like [3, 7, 17] apply an indirect method by converting the problem to edge points detection issue and point pair matching issue. He and Sun et al. [9] proposed an edge points detection strategy that develops a deep 3D edge points Hough voting neural network to learn the point-wise 3D offset and vote for 3D edge points. The 6DoF pose parameters are obtained by the least-squares fitting algorithm. Whereas, all these approaches need to sample huge number of seed points from the image. Hence, it is necessary to implement a secondary selection from seed points to optimize the training process.

In this work, we propose a dynamic keypoints selection algorithm that performs keypoints selection on seed points dense feature that consists of geometry information in point cloud and appearance information in RGB. Our key insight is that keypoints chosen from seed points can serve as complementary information during edge points detection procedure. We follow the pipeline proposed in FFB6D [15], which opens up new opportunities for feature fusion network and the 3D edge points based 6D pose estimation. However, it considers all seed points for 6D pose estimation. The background points have no contribution to the final pose prediction process. Hence, it is significant to set background points' weight in training and directly filter them after instance semantic in the pose estimation process. Filtering background points in this way is easier for the network to achieve the speed requirement of real-time tasks and the pose estimation performance is facilitated.

To fully evaluate our approach, we further conduct experiments on two benchmark datasets, the YCB-Video and LineMOD datasets. Experimental results show that our approach outperforms current state-of-the-art methods.

To summary, the main contributions of this work are:
- A novel but effective change based FFB6D for fully using scene information for 6DoF pose estimation from a single RGBD image.
- A simple but effective dynamic keypoints selection algorithm that leverages texture and geometry information of object mesh models
- State-of-the-art 6DoF pose estimation performance on the YCB-Video and LineMOD datasets.

The rest of the paper is organized as follows: Section 2 discusses the related work. Section 3, we explain the proposed method for 6DoF pose estimation. Section 4, we demonstrate the result of the proposed method and discuss the result obtained. Section 5 concludes the paper.

2. Related Work

**Pose Estimation with RGB Data.** Object pose estimation based on RGB data can be divided into three classes, local features matching, template matching and 2D-keypoints matching. Approaches [6, 8, 30, 34] apply machine learning or deep learning to use local features matching for 6DoF pose estimation, significantly improving the ability to resist object occlusion. Through finding the correspondence between mesh vertexes and image pixels, poses are recovered within Perspective-n-Point (PnP) manners. Though robust to occlusion, the demand for rich texture on objects limits the generalization of these approaches to textureless objects. The template matching method [11, 37] has the ability to handle textureless objects. Methods get the template of an object from different viewpoints and determine the location and orientation of the target by matching these templates against the input image. However, the drawback of methods is not robust to occlusions between objects. 2D keypoints-based method [3, 7, 13] detect 2D keypoints of objects from RGB image to build the 2D-3D correspondence for pose estimation. However, the loss of geometry information and the object occlusion limit the performance of these RGB-only methods.

**Pose Estimation with Point Clouds.** Given a 3D model of an object and a depth image, using camera intrinsic matrix transform depth image to point cloud, then the problem is formulated as aligning the two point clouds. The Iterative Closest Point (ICP) algorithm [2] is the most famous local refinement algorithm. The algorithm finds the correspondences between points and refines the pose estimation using the new correspondences given an initial pose estimation. However, without a close enough initial estimation, the algorithm may converge to a local optimum. Global registration methods do not assume an initial pose estimate as in [31] but computationally expensive, and the demand for 3D model quality limits the generalization of these methods. More and more deep learning methods using several point clouds only are motivated by the development of effective representation method of point cloud. These approaches feed point cloud into 3D ConvNets [20, 23, 27] or point cloud network [22, 28, 44] for geometry feature extraction to directly get the 6DoF pose estimation. However, depth sensors can not capture objects with reflective surfaces and sparsity point cloud limit the performance of object detection. Therefore, taking RGB images into account is necessary.

**Pose Estimation with RGBD Data.** When both depth images and RGB images are both available, they can be integrated to improve 6DoF pose estimation. Xiang and Schmidt et al. [19] propose PoseCNN in which the network directly regresses the initial pose from RGB images and uses ICP to refine it with the point cloud. However, they are not end-to-end optimizable and time-consuming. Instead, works like [10, 39] tackle the scene information by processing the RGB image and point cloud individually. Wang and Xu et al. [21] extract scene information from RGB images and point clouds using CNN and PCN individually and perform pixel-wise fusion for pose estimation. However, whether in the testing process or the training process, only one object is processed at one time. Moreover, He et al. [15] propose a full flow bidirectional fusion network and turn the pose estimation problem into an edge points matching problem. The 3D coordinate regression establishes correspondences between 3D scene edge points and 3D model edge points, from which 3D scene edge points are obtained by a deep Hough voting network and 6D pose can be computed by solving a least-squares problem. Nevertheless, since edge points and center points voting network need to select numerous points from the scene point cloud as input randomly, the network can not meet real-time requirements. In this work, we add a dynamic keypoints selected algorithm and a filtering background points algorithm into the network, which can fully utilize the scene information for pose estimation.

## 3. Proposed Method

Estimating the 6 DoF pose of a set of known objects present in an RGBD image is to estimate the rigid transformation matrix that transforms the object from its object world coordinate system to the camera world coordinate system. Without loss of generality, we represent 6DoF poses as a transformation matrix, $p \in SE(3)$ composed by a rotation $R \in SO(3)$ and a translation $t \in \mathbb{R}^3$.

### 3.1. Overview

To deal with this task, we develop an innovative approach based on a dynamic keypoints selection, as shown in Figure 1. The method is a two-stage pipeline followed by a pose parameters fitting module. Given an RGBD image as input, we first convert the depth image to a point cloud with camera intrinsic matrix. A fully convolutional network extracts the appearance information from RGB image, and a point cloud network extracts geometric information from point clouds. A full flow bidirectional fusion network is added into each layer during the feature extraction process that merges both appearance feature and geometry information. The extracted point-wise features are then fed into a keypoints selecting algorithm, a 3D edge points detection module, and an instance semantic segmentation module to obtain per-object 3d edge point in the single image. Finally, a least-square fitting algorithm is applied to the predicted edge point to recover 6DoF pose parameters, as in [15].

Unlike FFB6D, we apply a dynamic keypoints selection algorithm to reinforce the importance of the randomly selected seed points during training and a filtering background points algorithm to optimize the pose estimation process. More specifically, we take a foreground mask as input to filter the background points feature and apply a dynamic keypoints selection algorithm to choose the keypoints from the foreground in training. During 6DoF pose parameters estimation process, we transform the two-stage pipeline model to the one-stage and utilize the foreground mask obtained from the instance semantic segmentation process to filter out background points. Then, only foreground features are fed into the network to predict the translation offsets to edge points and the center point.

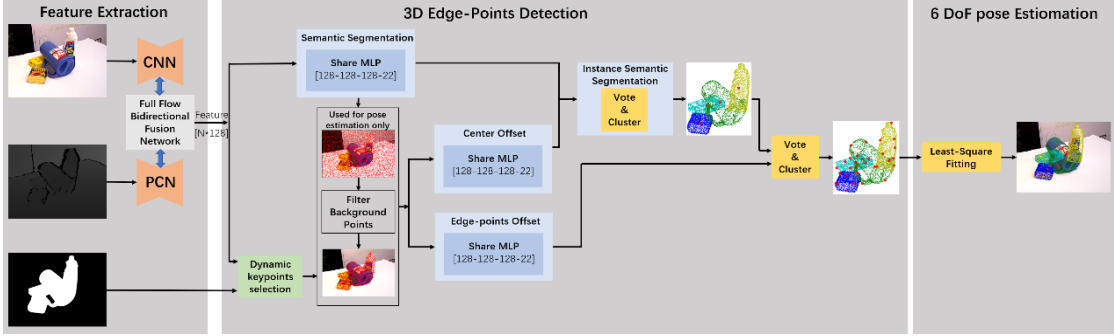

Figure 1. Overview of the method. The point-wise features are extracted from an RGBD image by the Feature Extraction module and then fed into the model to predict semantic labels and translation offsets to edge points, the center point of each point respectively and a dynamic keypoints selection algorithm to selected keypoints. A clustering algorithm is then applied to distinguish different objects and points on the same instance vote for their target edge points. A least-square fitting and background filtering algorithms are applied to the predicted edge points to estimate 6DoF pose parameters.

### 3.2. Dynamic Keypoints Selection Algorithm

It can be difficult to select the keypoints from the randomly selected points. Previous methods [25, 9] try to use the farthest point sampling (FPS) algorithm to choose the keypoints on the mesh. However, the algorithm only takes the Euclidean distance into account without distinctive texture and consumes time. To fully leverage objects texture and geometry information and the time efficiency, we propose a simple but effective 3D keypoints selection algorithm. To be special, we take the point-wise feature consisting of the appearance and geometry information as input to find the corresponding points of the maximum in the feature dimension. Then, we count the number of occurrences of corresponding points, sort the results and choose the first $k$ maximum points as the keypoints. By doing so, the selected keypoints algorithm is distinctive in texture and geometry and consumes a small enough time.

### 3.3. 3D Edge points Detecting Module

3D Edge points Detecting Module is used to detect the 3D edge points of each object. Significantly, the module predicts the point-wise Euclidean translation offset from randomly selected visible points to target edge points. These points and the corresponding predicted offsets then vote for the target edge points. Given a set of seed points $\{p_i\}_{i=1}^{N}$, a set of selected edge points $\{ep_j\}_{j=1}^{M}$ and a set of dynamic selected keypoints $\{kp_j\}_{j=1}^{K}$. The voted clustering algorithms gather edge points as follow [15]. To supervise the learning of the translation offset from the $i_{th}$ seed point to the $j_{th}$ edge point, we apply an L1 loss:

$$L_{edgepoints} = \frac{1}{N}\sum_{i=1}^{N}\sum_{j=1}^{M}\left\|of_i^j - of_i^{j*}\right\| * W(p_i)$$

$$where\ W(p_i) = \begin{cases} w_{keypoint} & p_i \in Keypoints \\ w_{background} & p_i \in background\ points \\ w_{others} & otherwise \end{cases} \quad (1)$$

Where $of_i^j$ denotes the translation offset from the $i_{th}$ seed point to the $j_{th}$ edge point. $of_i^{j*}$ is the ground truth translation offset; $M$ is the total number of selected target edge points; $N$ is the total number of seeds. $W(p_i)$ is a weight function equates to $w_{keypoint}$ only when point $p_i$ belongs to keypoints, equates to $w_{background}$ only when point $p_i$ belongs to background points, and equates to $w_{others}$ when point $p_i$ belongs to otherwise.

### 3.4. Instance semantic segmentation module

Our semantic segmentation network takes a pixel feature map as input and consists of two parts. The semantic segmentation module generates a semantic segmentation map, and the center offset model predicts the translation offsets to the center point. To be specific, given the point-wise extracted feature, the semantic segmentation module predicts the point-wise semantic labels. We supervise this module with Focal Loss:

$$L_{semantic} = -\alpha(1-q_i)^\gamma \log(q_i) \quad where \quad q_i = c_i \cdot l_i \quad (2)$$

With $\alpha$ the balance parameter, $\gamma$ the focusing parameter, $c_i$ the predicted confidence for the $i_{th}$ point that belongs to each class, and $l_i$ the one-hot representation of the ground true class label.

The center voting module takes in the point-wise feature but predicts the Euclidean translation offset $\Delta_{x_i}$ to the center of its instance. The learning of $\Delta_{x_i}$ is also supervised by an L1 loss:

$$L_{center} = \frac{1}{N} \sum_{i=1}^{N} \|\Delta x_i - \Delta x_i^*\| * W(p_i)$$

$$where\ W(p_i) = \begin{cases} w_{keypoint} & p_i \in Keypoints \\ w_{background} & p_i \in background\ points \\ w_{others} & otherwise \end{cases} \quad (3)$$

Where $N$ denotes the total number of seed points on the object surface and $\Delta x_i^*$ is the ground truth translation offset from seed $p_i$ to the instance center. $W(p_i)$ is a weight function equate to $w_{keypoint}$ only when point $p_i$ belongs to keypoints, equate to $w_{background}$ only when the point $p_i$ belongs to background points, and set $w_{others}$ when point $p_i$ belongs to otherwise.

### 3.5. Multi-task loss

We supervise the learning of module jointly with multi-tasks loss:

$$L_{multi-task} = \lambda_1 L_{edge-points} + \lambda_2 L_{center} + \lambda_3 L_{semantic} \quad (4)$$

Where $\lambda_1$, $\lambda_2$ and $\lambda_3$ are the weights for each task. Previous work [9] has proven that jointly training these tasks boosts the performance of each other.

### 3.6. 6D Object Pose Estimation

For each object instance, the 3D edge points voting module learns the point-wise offsets to the selected edge points that vote for them within a MeanShift [29] clustering manners. Given two points sets of an object, one from the $M$ detected edge points $\{ep_j\}_{j=1}^{M}$ in the camera coordinate system and another $\{ep_j'\}_{j=1}^{M}$ from their corresponding points in the object coordinate system, the 6DoF pose estimation module computes the pose parameters $(R,t)$ with a least-squares fitting algorithm [1], which finds $R$ and $t$ by minimizing the following square loss:

$$L_{least-squares} = \sum_{j=1}^{M} \|kp_j - (R \cdot kp'_j + t)\|^2 \quad (5)$$

Where $M$ is the number of selected edge points of an object.

## 4. Experiments

### 4.1. Datasets

To evaluate the performance of the proposed method, we evaluate our method on two 6DoF object pose estimation benchmark datasets.

YCB-Video [32] contains 92 RGBD videos in various indoor scenes that capture scenes of 21 YCB objects in varying shape and texture. The videos are annotated with 6D poses and instance semantic masks. We use the same training and testing set as prior work [9, 19, 21] to split the dataset into 80 videos for training and 2949 keyframes chosen from the rest 12 videos for testing. The same 80,000 synthetic images released by [19] are also taken for training.

LineMOD [11] consists of 13 low-textured objects in 13 videos. The challenge of this dataset is the texture-less objects, cluttered scenes, and varying lighting. It is widely adopted by both traditional approaches [41, 43] and recent learning-based methods [21, 42]. We split the training and testing set following previous works [19, 25] and add additional synthetic data released by [9, 15, 25] into our training set.

### 4.2. Evaluation Metrics

The average distance ADD and ADD-S metrics are adopted for thorough evaluation. The ADD metric is designed for asymmetric objects and the average distance is computed based on the mean point-pair distance between object vertexes transformed by the predicted 6D pose $[R,t]$ and the ground true pose $[R^*,t^*]$:

$$ADD = \frac{1}{q} \sum_{x \in o} \|(Rx+t) - (R^*x+t^*)\| \quad (6)$$

Where $x$ denotes a vertex of totally $q$ vertexes in object $o$. For symmetric objects, the ADD-S metric calculates the mean distance based on the closest point distance:

$$ADD-S = \frac{1}{q} \sum_{x_1 \in o} \min_{x_2 \in o} \|(Rx_1+t) - (R^*x_2+t^*)\| \quad (7)$$

In the YCB-Video dataset, we follow [9, 15, 19] and report the area under ADD-S and ADD(S) curve (AUC), which is obtained by varying the metrics distance threshold in evaluation. The ADD(S) is calculated similarly but computes ADD distance for asymmetric objects and ADD-S distance for symmetric objects. In the LineMoD dataset, we follow previous methods [11] and report the accuracy of ADD(S) distance of less than 10% of the objects diameter (ADD(S)-0.1d).

### 4.3. Implementation Details

**Network architecture.** The RGB image embedding network applies pre-trained ResNet34 [5] on ImageNet [4] as encoder and follows by a PSPNet [35] as decoder. We randomly sample $N = 12000$ points from depth images for point cloud feature extraction and use RandLA-Net [28] for point cloud feature representation learning. In each encoding and decoding layer of the two networks, a full flow bidirectional fusion network is applied to get the pixel-wise feature fusion. After the feature fusion process, each sampled seed point gets $f_i \in \mathbb{R}^C$ of $C$ dimension feature. These pixel-wise dense RGBD features are then fed into the instance semantic segmentation and the edge points offset learning modules consisted of shared MLPs.

**Optimization regularization.** Focal Loss [36] is applied in the semantic segmentation branch. L1 loss is applied in the 3D edge points offset and center point offset modules. To jointly optimize the three tasks, a multi-task loss set the weight $\lambda_1 = 3$, $\lambda_2 = \lambda_3 = 1$ in Formula 4. We sample $K = 25$ keypoints for each frame of RGBD image and set $w_{keypoint} = 2$, $w_{background} = 0$, $w_{others} = 1$ in Formula 1 and Formula 3.

**Edge points selection algorithm.** We apply SIFT-FPS edge points selection algorithm to select $M = 8$ target edge points out of them as in [15]. All the experiments are implemented in PyTorch with an NVIDIA 2080 GPU and Intel i9-9900k CPU 4.0 GHz processor.

### 4.4. Experiments on YCB-Video & LineMod Dataset

We compare our proposed method without any interactive refinement procedure with state-of-the-art (SOTA) methods. For a fair comparison, all the results of the methods are obtained by running source codes or pre-trained modules by the authors. Table 1 demonstrates the quantitative evaluation results for all the 21 objects in the YCB-Video Datasets compared with PoseCNN [19], DenseFusion [21], PVN3D [9], FFB6D [15]. As can be seen clearly, our method advances the best performance in all the metrics on the YCB-Video Datasets. Especially on the ADD(S), we outperform the second-best result by 1.4%. Table 2 shows the evaluation results on the LineMOD dataset compared with PoseCNN [19, 25, 26, 21, 9, 15]. For the LineMOD dataset, our proposed method also performs better than other methods. Figure 2 shows visual comparison results of our method on the YCB-Video dataset with a recent state-of-the-art method [15]. Points on different meshes are projected back to the image after being transformed by the predicted pose. Here, we draw the rectangular region of the object for better visualization. It can be intuitively observed that our method successfully predicts the 6 DoF pose of all items in the scene and is robust in heavily occluded scenes.

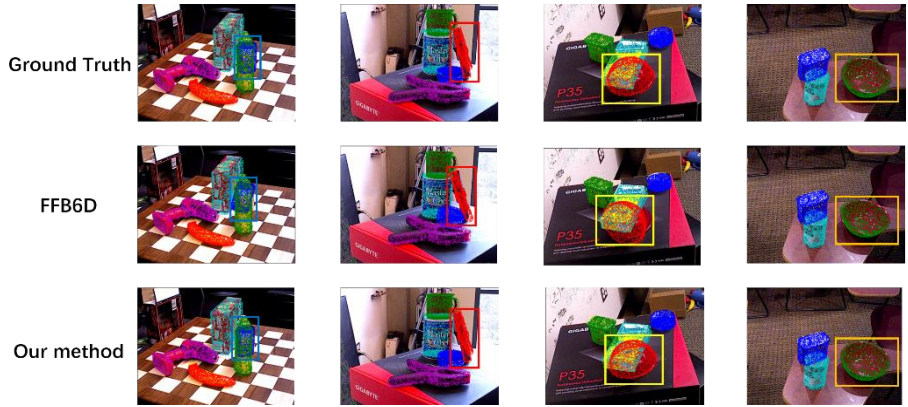

Figure 2. Qualitative results on the YCB-Video dataset. Points on different meshes in the same scene are in different colors. We compare our method without iterative refinement to FFB6D.

|  | PoseCNN[19] | | DenseFusion[21] | | PVN3D[9] | | FFB6D[15] | | Ours | |
|---|---|---|---|---|---|---|---|---|---|---|
| Object | ADDS | ADD(S) | ADDS | ADD(S) | ADDS | ADD(S) | ADDS | ADD(S) | ADD | ADD(S) |
| Master_chef_can | 83.9 | 50.2 | 95.3 | 70.7 | 96.0 | 80.5 | 96.3 | 80.6 | **96.4** | **81.5** |
| Craker_box | 76.9 | 53.1 | 92.5 | 86.9 | 96.1 | 94.8 | 96.3 | 94.6 | **96.5** | **95.1** |
| Sugar_box | 84.2 | 68.4 | 95.1 | 90.8 | 97.4 | 96.3 | 97.6 | 96.6 | **97.8** | **96.9** |
| Tomato_soup_can | 81.0 | 66.2 | 93.8 | 84.7 | 96.2 | 88.5 | 95.6 | **89.6** | 95.9 | 88.3 |
| Mustard_bottle | 90.4 | 81.0 | 95.8 | 90.9 | 97.5 | 96.2 | 97.8 | 97.0 | **98.2** | **97.7** |
| Tuna_fish_can | 88.0 | 70.7 | 95.7 | 79.6 | 96.0 | 89.3 | 96.8 | 88.9 | **97.3** | **92.1** |
| Pudding_box | 79.1 | 62.7 | 94.3 | 89.3 | 97.1 | **95.7** | 97.1 | 94.6 | 96.4 | 93.4 |

| | | | | | | | | | |
|---|---|---|---|---|---|---|---|---|---|
| Gelatin_box | 87.2 | 75.2 | 97.2 | 95.8 | 97.7 | 96.1 | **98.1** | **96.9** | 97.8 | 95.9 |
| Potted_meat_can | 78.5 | 59.5 | 89.3 | 79.6 | 93.3 | 88.6 | **94.7** | **88.1** | 92.9 | 90.3 |
| Banana | 86.0 | 72.3 | 90.0 | 76.7 | 96.6 | 93.7 | 97.2 | **94.9** | **97.3** | 94.4 |
| Picher_base | 77.0 | 53.3 | 93.6 | 87.1 | 97.4 | 96.5 | 97.6 | 96.9 | **97.8** | **97.2** |
| Bleach_cleanser | 71.6 | 50.3 | 94.4 | 87.5 | 96.0 | 93.2 | **96.8** | **94.8** | 96.5 | 93.8 |
| **Bowl** | 69.6 | 69.6 | 86.0 | 86.0 | 90.2 | 90.2 | 96.3 | 96.3 | **96.5** | **96.5** |
| Mug | 78.2 | 58.5 | 95.3 | 83.8 | 97.6 | 95.4 | 97.3 | 94.2 | **97.6** | 95.6 |
| Power_drill | 72.7 | 55.3 | 92.1 | 83.7 | 96.7 | 95.1 | 97.2 | 95.9 | **97.3** | 96.3 |
| **Wood_block** | 64.3 | 64.3 | 89.5 | 89.5 | 90.4 | 90.4 | 92.6 | 92.6 | **93.6** | **93.6** |
| Scissors | 56.9 | 35.8 | 90.1 | 77.4 | 96.7 | 92.7 | 97.7 | 95.7 | **98.1** | 97.1 |
| Large_marker | 71.7 | 58.3 | 95.1 | 89.1 | 96.7 | 91.8 | 96.6 | 89.1 | **97.0** | 90.2 |
| **large_clamp** | 50.2 | 50.2 | 71.5 | 71.5 | 93.6 | 93.6 | 96.8 | 96.8 | **97.1** | **97.1** |
| **extra_ clamp** | 44.1 | 44.1 | 70.2 | 70.2 | 88.4 | 88.4 | 96.0 | 96.0 | **96.1** | **96.1** |
| **foam_brick** | 88.0 | 88.0 | 92.2 | 92.2 | 96.8 | 96.8 | 97.3 | 97.3 | **97.9** | **97.9** |
| ALL | 75.8 | 59.9 | 91.2 | 82.9 | 95.5 | 91.8 | 96.6 | 92.7 | **96.9** | 94.1 |

Table 1. Quantitative evaluation of 6DoF pose without iterative refinement on YCB-video Dataset. The ADD-S and ADD(S) AUC are reported. Objects with bold names are symmetric.

| | RGB | | RGB-D | | | | |
|---|---|---|---|---|---|---|---|
| Object | PoseCNN[19] | PVNe[25]t | Point-Fusion[26] | Dense-Fusion[21] | PVN3D[9] | FFB6D[15] | Ours |
| Ape | 77.0 | 43.6 | 70.4 | 92.3 | 97.3 | **99.1** | 98.8 |
| Benchvise | 97.5 | 99.9 | 80.7 | 93.2 | 99.7 | **100.0** | **100.0** |
| Camera | 93.5 | 86.9 | 60.8 | 94.4 | 99.6 | **100.0** | **100.0** |
| Can | 96.5 | 95.5 | 61.1 | 93.1 | 99.5 | **100.0** | **100.0** |
| Cat | 82.1 | 79.3 | 79.1 | 96.5 | 99.8 | 99.5 | **99.9** |
| Driller | 95.0 | 96.4 | 47.3 | 87.0 | 99.3 | 99.8 | **100.0** |
| Duck | 77.7 | 52.6 | 63.0 | 92.3 | 98.2 | 98.4 | **98.5** |
| **Eggbox** | 97.1 | 99.2 | **99.9** | 99.8 | 99.8 | 99.5 | 99.8 |
| **Glue** | 99.4 | 95.7 | 99.3 | **100.0** | **100.0** | **100.0** | **100.0** |
| Holepucher | 52.8 | 82.0 | 71.8 | 92.1 | 99.9 | **100.0** | **100.0** |
| Iron | 98.3 | 98.9 | 83.2 | 97.0 | 99.7 | 99.9 | **100.0** |
| Lamp | 97.5 | 99.3 | 62.3 | 95.3 | 99.8 | 99.9 | **100.0** |
| Phone | 87.7 | 92.4 | 68.8 | 92.8 | 99.5 | **100.0** | 99.9 |
| Mean | 88.6 | 86.3 | 73.7 | 94.3 | 99.4 | 99.7 | 99.8 |

Table 2. Quantitative evaluation of 6D Pose on ADD(S) metric on the LineMOD dataset. Symmetric objects are in bold.

### 4.5. Ablation Study

In this subsection, we explore the influence of our design choices for 6D pose estimation and discuss their effect.

**Effect of filtering background points.** In this part, we conduct the experiment to evaluate the effect of filtering background points algorithm. Compared with using all seed points that much time is consumed in calculating the distance between background points and edge points or center points, our method fully utilizes the instance semantic segmentation result to filter background points for pose estimation. We also evaluate the effect of the proposed filtering background points algorithm by applying it in FFB6D. Shown in Table 3, the proposed algorithm significantly outperforms both time efficiency and accuracy by a large margin.

| FBP | method | ADD-S | ADD(s) | Run-time(ms/frame) |
|---|---|---|---|---|
| × | FFB6D | 96.6 | 92.7 | 142 |
| √ | FFB6D | 96.7 | 93.1 | 100 |
| × | Ours | 96.7 | 93.8 | 145 |
| √ | Ours | 96.9 | 94.1 | 105 |

Table 3: Effect of filtering background points on the YCB-Video Dataset of FFB6D and our method. FBP means the filtering background points algorithm.

**Effect of dynamic keypoints selection algorithm.** In Table 4, we study the effect of different keypoints selection algorithms. Compared with FPS that only considers the mutual distance between keypoints and edge points, our dynamic keypoints selection takes both object texture and geometry information into account and saves time. In order to prove the effect of the method, we project the 3D keypoints and 3D edge points back to 2D by the camera intrinsic parameters with different training time on one scene in Figure 3. Our keypoints are close to the geometric and texture edge of the object. Therefore, these points results have more influence than other points.

|  | FPS8 | DKS-10 | DKS-25 | DKS-50 |
|---|---|---|---|---|
| KP err.(cm) | 1.5 | 1.0 | 0.8 | 0.9 |
| ADD-S | 96.6 | 96.5 | 97.0 | 96.9 |
| ADD(S) | 92.7 | 92.9 | 94.1 | 93.1 |

Table 4: Effect of dynamic keypoints Selection algorithm. DKS means the proposed dynamic keypoints selection algorithm.

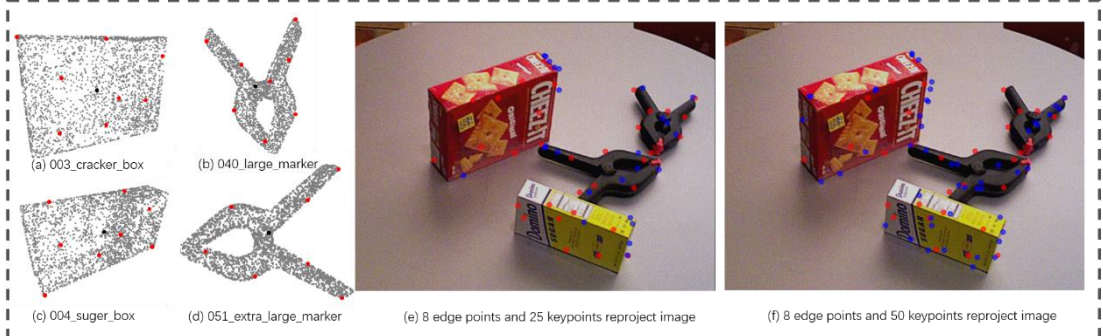

Figure 3: 3D keypoints and 3D edge points during training. (a-d) parts are the edge points showen in 3D model. (e-f) parts are the projected edge points (red) and the dynamic keypoints (blue).

**Effect of Multi-task learning to enhance semantic segmentation result.** In this part, we explore how jointly trained with edge points, and center offset improves the accuracy of semantic segmentation. In Table 5, we explore how edge points and center points offset learning improve the semantic module. Point mean intersection over union (mIoU) is used as evaluation metric. We compare our module with other methods based RGB or RGBD [9, 19, 45]. With extra depth information and full flow bidirectional fusion network, our jointly trained model surpasses all other approaches by a small margin. Meanwhile, we find jointly trained model improves the accuracy of object foreground mask estimation. Some qualitative results are shown in Figure 4.

|  | RGB | | RGBD | |
|---|---|---|---|---|
|  | PoseCNN [19] | Mask R-CNN[45] | PVN3D[9] | Ours |
| mIoU(%) | 91.5 | 95.6 | 95.8 | 96.1 |

Table 5: Instance semantic segmentation results (mIoU(%)) of different methods on the YCB-Video dataset.

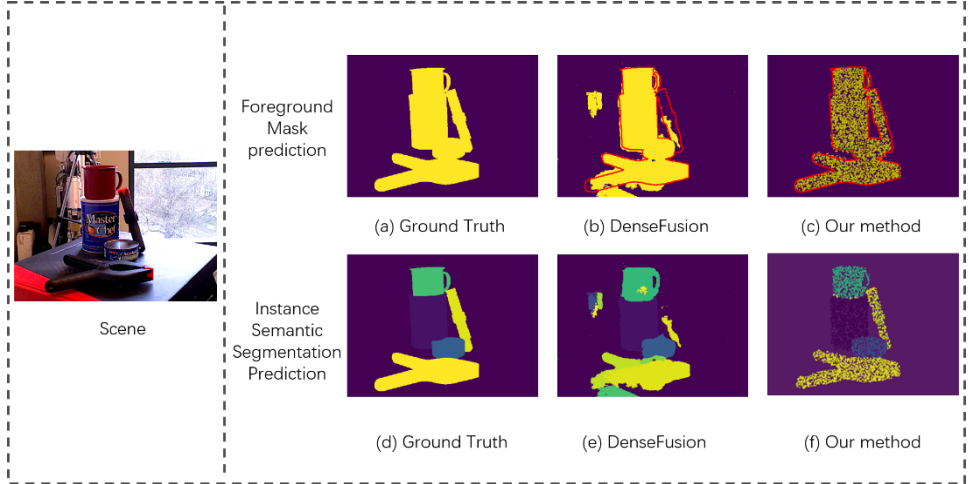

Figure 4: Effect of Multi-task learning to enhance semantic segmentation result. The top of the figure is the ground truth and the different method prediction results of foreground mask prediction. (a) shows the ground truth label. In (b)-(c) show our method can distinguish it well. The below figure shows the ground truth (d) and the different method prediction results of instance semantic segmentation results (e)-(f).

**Model parameters and time efficiency.** In Table 6, we report the parameters and run-time breakdown of our method. Compared to FFB6D [15] and PVN3D [9], our keypoints selection and filtering background points methods achieve better performance with fewer parameters and is 1.4 times faster.

|  | Parameters | Run-time(ms/frame) | | |
| --- | --- | --- | --- | --- |
|  |  | NF | PE | ALL |
| FFB6D[15] | 39.2M | 170 | 90 | 260 |
| PVN3D[9] | 33.8M | 55 | 87 | 142 |
| Ours | 33.4M | 35 | 70 | 105 |

Table 6: Run-time breakdown on the YCB-Video dataset. NF: Network Forward; PE: Pose Estimation.

## 5. Conclusion

To obtain accurate 6DoF pose estimation from a single RGBD image, we introduce a simple but effective keypoints selection algorithm in this paper, which leverages texture and geometry information of objects to simplify keypoints for optimizing the network training process. Additionally, we design a filtering background points algorithm for accelerating 6DoF pose estimation process. Benefiting from the proposed algorithms, our network can satisfy the demands of fast inference simultaneously and accurate pose estimation. Besides, experiments show that our proposed approach outperforms the existing state-of-the-art model on two benchmark datasets, with advantages in time efficiency and boundary accuracy. Moreover, we believe the proposed dynamic keypoints select methods can generalize to more applications built on RGBD or only depth images, such as 3D object detection, and expect to see more future research along this line.

**Declaration of Competing Interest**

The authors declare that they have no known competing financial interests or personal relationships that could have appeared to influence the work reported in this paper.

**Acknowledgement**

This work is partially supported by the National Natural Science Foundation of China (No. 51975402), the Basic Innovation Team Program of China North Industries Group Corporation Limited (No. 2017CX031).